\documentclass[conference]{IEEEtran}
\usepackage{amsmath}
\usepackage{lipsum}
\usepackage{multicol}
\usepackage[pdftex]{graphicx}
\usepackage{graphicx}
\usepackage[export]{adjustbox}

\usepackage{float}
\floatstyle{plaintop}
\restylefloat{table}
\usepackage{placeins}

\ifCLASSINFOpdf
\else
\fi

\begin{document}

%

\title{SAFS: A Deep Feature Selection Approach for Precision Medicine}

\author{\IEEEauthorblockN{Milad Zafar Nezhad$^{a}$, Dongxiao Zhu$^{b,*}$, Xiangrui Li$^{b}$, Kai Yang$^{a}$, Phillip Levy$^{c}$}\\
\IEEEauthorblockA{Department of Industrial and Systems Engineering,
Wayne State University$^{a}$\\Department of Computer Science, Wayne State University$^{b}$\\Department of Emergency Medicine and Cardiovascular Research Institute, Medical School, Wayne State University$^{c}$\\
Corresponding author$^{*}$, E-mail addresses: dzhu@wayne.edu}
\and
}

\maketitle

\begin{abstract}

In this paper, we propose a new deep feature selection method based on deep architecture. Our method uses stacked auto-encoders for feature representation in higher-level abstraction. We developed and applied a novel feature learning approach to a specific precision medicine problem, which focuses on assessing and prioritizing risk factors for hypertension (HTN) in a vulnerable demographic subgroup (African-American). Our approach is to use deep learning to identify significant risk factors affecting left ventricular mass indexed to body surface area (LVMI) as an indicator of heart damage risk. The results show that our feature learning and representation approach leads to better results in comparison with others.   


\end{abstract}


%
\IEEEpeerreviewmaketitle

\section{Introduction}
The explosive increase of Electronic Medical Records (EMR) provides many opportunities to carry out data science research by applying data mining and machine learning tools and techniques. EMR contains massive and a wide range of information on patients concerning different aspects of healthcare, such as patient conditions, diagnostic tests, labs, imaging exams, genomics, proteomics, treatments, outcomes or claims, financial records \cite{IEEEhowto:1}. Particularly, the extensive and powerful patient-centered data enables data scientists and medical researchers to conduct their research in the field of personalized (precision) medicine. Personalized medicine is defined as \cite{IEEEhowto:2}: ``the use of combined knowledge (genetic or otherwise) about a person to predict disease susceptibility, disease prognosis, or treatment response and thereby improve that person’s health." In other words, the goal of precision medicine or personalized healthcare is to provide “the right treatment to the right patient at the right time”. Personalized medicine is a multi-disciplinary area that combines data science tools and statistics techniques with medical knowledge to develop tailor-made treatment, prevention and intervention plans for individual patients.

In this study we focus on a vulnerable demographic subgroup (African-American) at high-risk for hypertension (HTN), poor blood pressure control and consequently, adverse pressure-related cardiovascular complications. We use left ventricular mass indexed to body surface area (LVMI) as an indicator of heart damage risk. The ability to reduce LVMI would lead to an improvement in hypertension-related cardiovascular outcomes, which, in turn, would further diminish the excess risk of cardiovascular disease complications in African-Americans. 

Based on individual clinical data with many features, our objective is to identify and prioritize personalized features to control and predict the amount of LVMI toward decreasing the risk of heart disease. First, we focus on selecting significant features among many features such as demographic characteristics, previous medical history, patient medical condition, laboratory test result, and Cardiovascular Magnetic Resonance (CMR) results that are used to determine LVMI and its reduction with treatment over time. Second, after features selection, we apply supervised machine learning methods to predict the amount of LVMI as a regression model. Third, this prediction method can be implemented as a decision support system (DSS) to assist medical doctors in controlling the amount of LVMI based on significant risk factors. 

\begin{figure*}
 \centering
 \includegraphics[scale= 0.22]{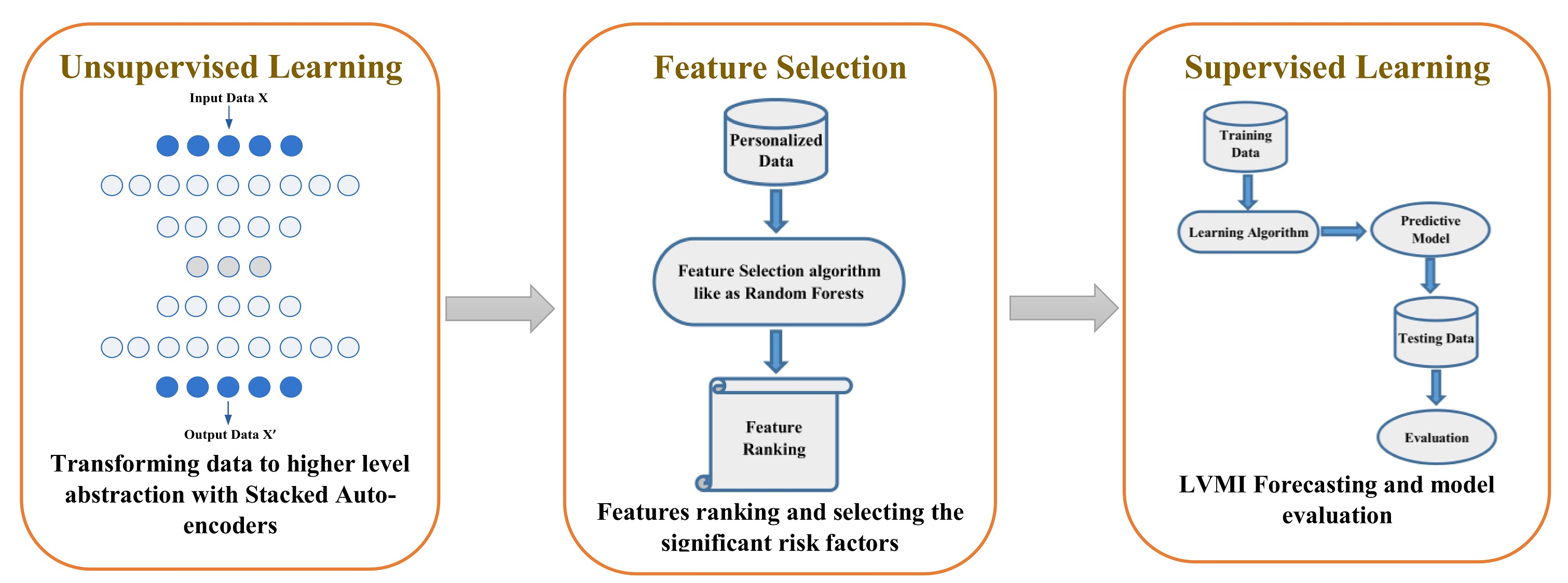}
 \caption{An Illustration of the Three Consecutive Steps for our SAFS Approach}\label{Fig1}
 \end{figure*}

Figure \ref{Fig1} illustrates our approach in three consecutive steps; this approach is an integrated feature selection model that applies unsupervised learning for producing higher-level abstraction of input data and then using them as an input of a feature selection method for ranking important features. In the final step, supervised leaning method is implemented for forecasting LVMI and model evaluation. Based on the model evaluation results, steps $2$ and $3$ are applied iteratively to finalize the most important features (risk factors).

In this paper, we develop a new feature selection and representation method using a deep learning algorithm for personalized medicine. In our method, we apply stacked auto-encoders \cite{IEEEhowto:10} as a deep architecture for feature abstraction at higher levels. To our knowledge, it is one of the first methods that use stacked auto-encoders for feature learning and selection, the LVMI prediction approach developed in this research is also a new model of risk forecasting for cardiovascular and other health conditions.

The rest of this paper is organized as follows. Section II reviews the related works in deep feature selection methods. Section III explains deep learning overview and stacked auto-encoders. Section IV describes the developed feature selection method and prediction approach. Section V reports the personalized data and implementation results and finally section VI discusses about the results and conclusion. 

\section{Related Works}
Deep learning, including feature representation and predictive modeling, has been researched and applied in a number of areas, such as computer vision, remote sensing, natural language processing and bioinformatics. The main reasons accounting for the extensive applications are improved prediction accuracy, capability of modeling processes of complex systems, and generating high-level representation of features and higher robustness modeling [12]. In this literature review, we focus on papers that develop feature selection models using deep learning algorithms. Among many related applications, few researchers have carried out feature selection using deep learning. 

Zhou et al. \cite{IEEEhowto:11} used deep feature selection methods in natural language processing and sentiment classification. They proposed a novel semi-supervised learning algorithm called active deep network (ADN) by using restricted Boltzmann machines (RBMs). They applied unsupervised learning based on labeled reviews and abundant unlabeled reviews and then used active learning to identify and select reviews that should be labeled as training data. Also in \cite{IEEEhowto:13}, a semi-supervised learning algorithm based on deep belief network, called DBNFS, was proposed for feature selection. Authors applied the method for solving a sentiment classification problem to reduce high dimension of noisy vocabularies. In their method, several hidden layers have been replaced in DBNs, which are computationally expensive with the filter-based feature selection technique using chi-squared tests. 

Deep feature selection was also applied in a Remote Sensing Scene classification/recognition problem. Zou et al. \cite{IEEEhowto:14} developed a new feature selection method by using deep learning that formulates the feature selection problem as a feature reconstruction problem. Their method is implemented as an iterative algorithm, which is based on DBN in an unsupervised learning way to train the inquired reconstruction weights. Finally features with small reconstruction errors were selected as input features for classification. 

In biomedical and biomedicine research, deep learning algorithms and tools have been applied in different areas especially in precision medicine and personalized healthcare by using biological and genomics data \cite{IEEEhowto:6}. 

Li et al. \cite{IEEEhowto:12} developed a deep feature selection (DFS) model for selecting input features in a deep neural network for multi-class data. They used elastic net to add a sparse one-to-one linear layer between the input layer and the first hidden layer of a multi-layer perception and select most important features according to their weights in the input layer given after training. They then applied their DFS model to solve the problem of enhancer-promoter interaction using genomics data. The method that they developed using elastic net is a new shrinkage approach that is flexible to be applied in different deep learning architectures. In terms of performance, their method did not outperform random forest in the experimental implementation that would be considered as a major weakness. 

Another study in the biomedical area \cite{IEEEhowto:15} used DBN and unsupervised active learning to propose a new multi-level feature selection method for selecting genes/miRNAs based on expression profiles. The strength of the proposed approach is better performance in comparison with several feature selection methods. However, for high dimensional features it does not outperform some other methods such as relief algorithm. 

In \cite{IEEEhowto:16} authors showed how unsupervised feature learning approach can be applied for cancer type classification from gene expression data. They proposed a two-phase model for feature engineering and then classification. In the first phase, Principal Component Analysis (PCA) technique was used for dimension reduction of the feature space and in the second phase, a deep sparse encoding was implemented on new data to obtain high-level abstractions for cancer diagnosis and classification. While their approach used stacked auto-encoders for representation learning, they focused on using deep feature extraction to achieve a better accuracy in cancer type prediction instead of finding significant factors in cancer diagnosis. Also in similar research \cite{IEEEhowto:17} focused on early detection of lung cancer, authors used deep learning for features extraction and applied a binary decision tree as a classifier to create a computer aided diagnosis (CAD) system that can automatically recognize and analyze lung nodules in CT images. They just implemented an auto-encoder for data representation and probably using stacked auto-encoders leads to higher accuracy. 

In sum, feature selection based on deep learning techniques were used in a wide range of areas as reviewed above, which is evident by an increasing amount of related studies that have been developed in recent years (after 2013). Particularly, many studies in biomedical applications used or developed deep learning methods for feature selection or feature extraction to reduce the dimensionality of data with many features that increase the performance of prediction and classification tasks.   

Precision medicine has emerged as a data-rich area with millions of patients being prospectively enrolled in the Precision Medicine Initiative (PMI). The data is big and noisy thus demanding for more sophisticated data science approaches. Therefore it can be concluded that there are ample opportunities to develop and apply deep learning for feature selection in the new data-rich areas, such as precision medicine. 

From what we discussed, most related research works took the advantages of representation learning with deep architecture. In many of them, DBN was applied for data transforming and there lack comprehensive and systematic comparisons of performance among different deep architecture. In our approach we will use stacked auto-encoders with carefully selected architectures for feature selection that can overcome this problem. 

\section{DEEP LEARNING AND STACKED AUTO-ENCODERS}
\subsection{Introduction  to deep learning and its applications}
Deep Learning or Deep Machine Learning is a kind of machine learning algorithms that model input data to higher-level abstraction by using a deep architecture with many hidden layers composed of linear and non-linear transformations \cite{IEEEhowto:5}\cite{IEEEhowto:7}\cite{IEEEhowto:19}. 


In a simple definition, deep learning means using a neural network with several layers of nodes between input and output and deep architectures that are composed of multiple levels of non-linear operations, such as neural nets with many hidden layers. There are three different types of deep architectures: 1) Feed Forward architectures (multilayer neural nets, convolutional nets), 2) Feed Back architectures (De-convolutional Nets) and 3) Bi-Directional architectures (Deep Boltzmann Machines, Stacked Auto-Encoders) \cite{IEEEhowto:5}.

Choice of data representation or feature representation plays a significant role in success of machine learning algorithms \cite{IEEEhowto:3}. For this reason, many efforts in developing machine learning algorithms focus on designing preprocessing mechanisms and data transformations for representation learning that would enable more efficient machine learning algorithms \cite{IEEEhowto:3}.

Deep learning applications encompass many domains. The major ones are Speech Recognition, Visual Object Recognition, Object Detection (Face Detection) and Bioinformatics or Biomedicine applications such as drug discovery and genomics \cite{IEEEhowto:7}. In biomedical and health science, increases in technological development, information systems and research equipment have generated a large amount of data with high features and dimensions. Since deep learning outperformed some other methods such as PCA or singular value decomposition in handling high-dimensional biomedical data, it has strong potential for dimension reduction and feature representation in biomedical and clinical research \cite{IEEEhowto:6}.

Deep Learning algorithms and tools have not been applied in biomedical and health science areas extensively. Deep supervised, unsupervised, and reinforcement learning methods can be applied in many complex biological and health personalized data and could play an important role in discovering and development of personalized healthcare and medicine \cite{IEEEhowto:6}. 

\subsection{Introduction to stacked auto-encoders}
Among different kinds of deep architectures mentioned in previous section, four architectures are more popular in biomedical data analysis [6]. 1) The Convolutional Neural Network (CNN), which usually consists of one or more convolutional layers with sub-sampling layers and followed by fully connected layers as in a standard deep neural network. 2) Restricted Boltzmann Machine (RBM), which is comprised of one visible layer and one layer of hidden units. 3) Deep Belief Network (DBN), which is a generative graphical model consisted of multiple layers of hidden variables with connections just between layers (not units within layers). 4) Stacked auto-encoders, which is a neural network including multiple layers of sparse auto-encoders [5][8]. 

\begin{figure}[H]
	\centering
	\includegraphics[scale= 0.2]{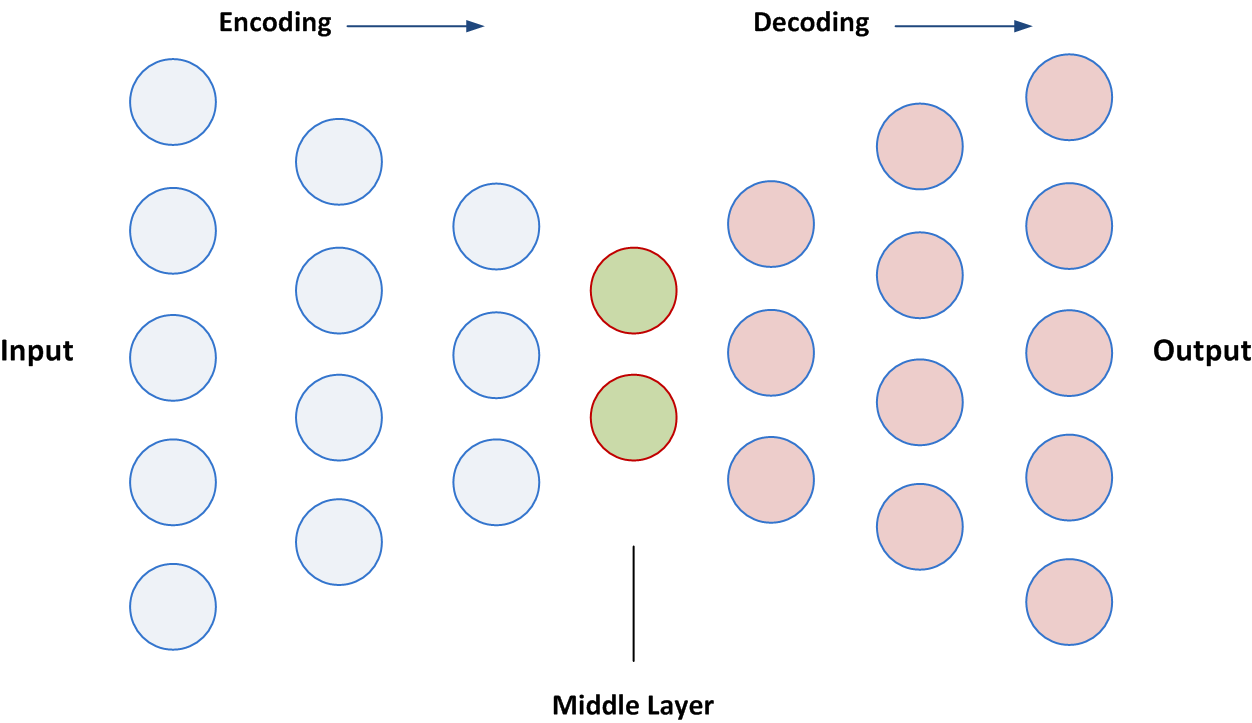}
	\caption{Stacked Auto-Encoders}\label{Fig2}
\end{figure}

Training deep neural networks with multiple hidden layers is known to be challenging. Standard approach for learning neural network applies gradient-based optimization with back-propagation by initializing random weights in network concludes poor training results empirically when there are three or more hidden layers in deep network [4]. Hinton et al in [18] developed a greedy layer-wise unsupervised learning algorithm for training DBN parameters by using a RBM in the top of deep architecture. Bengio et al. [10] used greedy layer-wise unsupervised learning to train deep neural network when layer building block is an auto-encoder instead of the RBM. These two greedy layer-wise algorithms are the major algorithms to train a deep architecture based on DBN and Stacked auto-encoders. In this study we use stacked auto-encoders for data representation learning. Stacked auto-encoders shown in Figure \ref{Fig2} is constructed by stacking multiple layers of auto-encoder. 

An auto-encoder is trained to reconstruct its own inputs by encoding and decoding processes. Let us define $w^{(k,l)}$, $w^{(k,2)}$, $b^{(k,l)}$, $b^{(k,2)}$ as the parameters of $k^{th}$ auto-encoder for weights in encoding and decoding process respectively. Encoding step of each layer is a forward process and mathematically described as below: 

\begin{flalign}\label{eq1}
&a^{(k)}= f(z^{\left( k\right) }),&\\
&z^{(k+1)}= w^{\left( k,1\right) }a^{\left( k\right) }+b^{\left( k,1\right) }.&
\end{flalign}

In equation \ref{eq1}, $f(x)$ is an activation function for transforming data. If $n$ represents the location of middle layer in stacked auto-encoders, the decoding step is to apply the decoding stack of each auto-encoder as following \cite{IEEEhowto:9}:

\begin{flalign}
&a^{(n+k)}= f(z^{\left(n+k\right) }),&\\
&z^{(n+k+1)}= w^{\left(n+k,2\right) }a^{\left(n+k\right) }+b^{\left(n+k,2\right) }.&
\end{flalign}

The training algorithm for obtaining parameters of stacked auto-encoders is based on greedy layer-wise strategy \cite{IEEEhowto:10}. It means that each auto-encoder should be trained by encoding and decoding process one-by-one. By training this deep architecture, $a^{(n)}$ (the middle layer) gives the highest representation of the input \cite{IEEEhowto:9}. In the simplest case, when an auto-encoder with sigmoid activation function has only one hidden layer and takes input $x$, the output of encoding process will be:
\begin{flalign}
&z= \mathrm{Sigmoid}_1(wx+b).&
\end{flalign}

Therefore $z$ is the vector of transformed input in the middle layer. In the second step (decoding process), $z$ is transformed into the reconstruction $x^{'}$, i.e., 
\begin{flalign}
&x^{'}= \mathrm{Sigmoid}_2(W^{'}z+b^{'}).&
\end{flalign}

Finally, auto-encoder is trained by minimizing the reconstruction errors as below:
\begin{flalign}
&\mathrm{Loss}(x, x^{'})= \Vert x- x^{'}\Vert = &\notag\\ 
& \Vert x- \mathrm{Sigmoid}_2(W^{'} (\mathrm{Sigmoid}_1(Wx+b))+b^{'}) \Vert.&
\end{flalign}

Auto-Encoder with rich representation is an appropriate deep learning architecture to implement on biomedical data \cite{IEEEhowto:6}. In the next section we will use stacked auto-encoders for feature learning and selection. 

\section{METHODOLOGY}

The method developed in this study is an integrated feature selection method with stacked auto-encoders, which is called SAFS. The work flow of our approach is shown in Figure \ref{Fig3}, and it includes three major components as follows:

\begin{figure}[H]
 \centering
  \includegraphics[scale= 0.18]{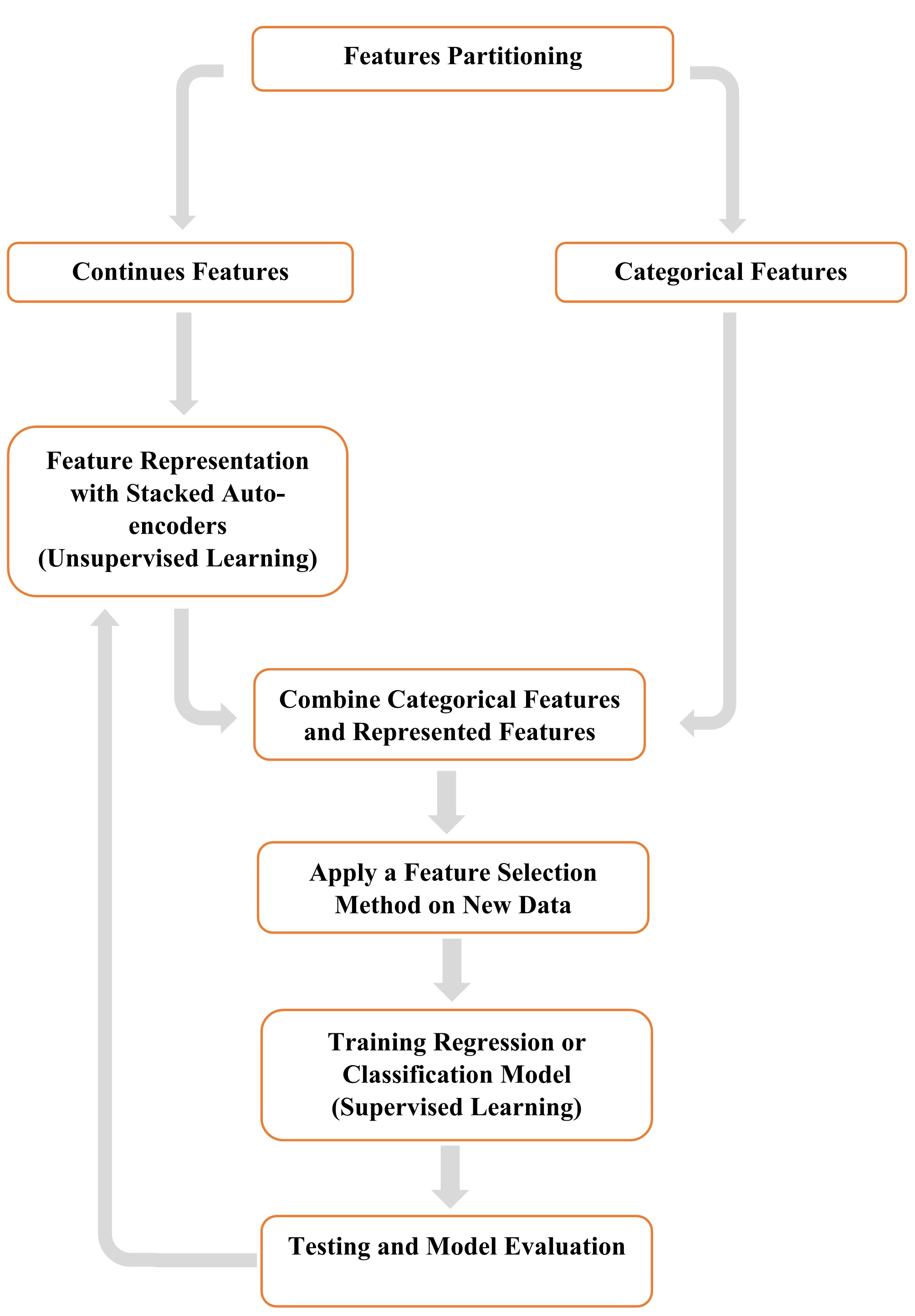}
 \caption{The Proposed SAFS Workflow}\label{Fig3}
\end{figure}

\subsection{Features Partitioning}

In the first step, we partition categorical features from continuous features when they co-exist in a data set. Here we focus on representation learning on continuous features since categorical features are often used as dummy variables in linear model based machine learning methods.  

\subsection{Features Representation by Stacked Auto-Encoders}

\begin{figure}[H]
 \centering
  \includegraphics[scale= 0.3]{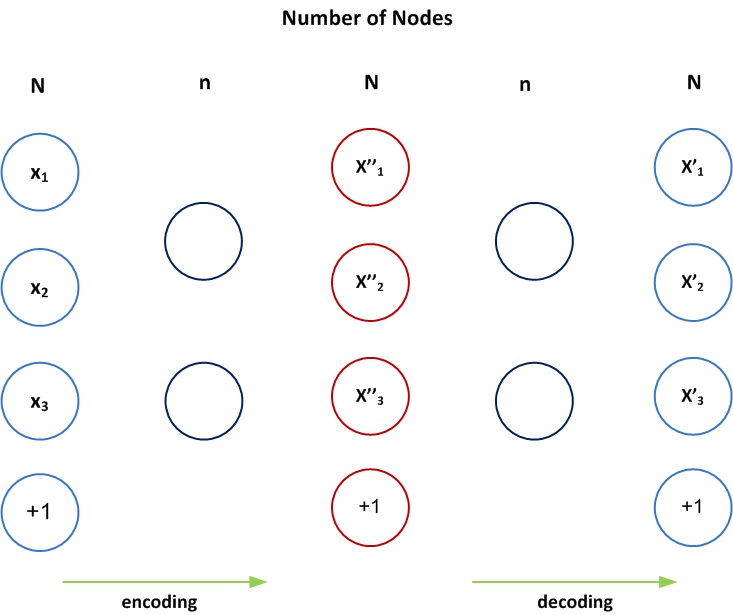}
 \caption{A Deep Architecture for SAE}\label{Fig4}
\end{figure}

The second step is unsupervised learning section. Continuous features are represented in higher-level abstraction by deep stacked auto-encoders. As mentioned in Section III stacked auto-encoders is one of the most popular deep architectures for representation learning especially for biomedical data. The deep architecture of stacked auto-encoders considered with $5$ layers as shown in Figure \ref{Fig4}. 

In this deep architecture, $N$ is assumed as the number of continuous features and $n$ is a parameter. The middle hidden layer has $N$ nodes, same as input and output layers, and the other two hidden layers have $n$ nodes as a variant. The represented data is given from middle layer and $n$ is determined in an iterative process between unsupervised learning step and supervised learning step. 

\subsection{Supervised Learning}

In the third step, the represented continuous features are combined with categorical features and then supervised learning is applied on new dataset. It starts with feature selection, which can use any feature selection method such as random forests. Selected important features from a main dataset are entered in a supervised classifier for regression or classification and after training step. Model are then evaluated by some specific measures. If the results meet the criteria then selected features considered as significant variables, if not, the model tries different architecture for stacked auto-encoders by changing the number of nodes ($n$) in two hidden layers. This iterative process is continued until the model converges to some stop criteria or fixed number of iterations. 

Our presented approach using deep learning for unsupervised non-linear sparse feature learning leads to a construction of higher level features for general feature selection methods and prediction tasks. 

\section{IMPLEMENTATION FOR PERSONALIZED MEDICINE }
In this section we implement our methodology and demonstrate its performance using a precision medicine case to find significant risk factors. We demonstrate that SAFS approach selects better features in comparison when we apply feature selection without using represented data.

The data used in this paper are derived from a subgroup of African-Americans with hypertension and poor blood pressure control who have high risk of cardiovascular disease. Data are obtained from patients who were enrolled in the emergency department of Detroit Receiving Hospital. Among many features (more than 700) including demographic characteristics, previous medical history, patient medical condition, laboratory test results, and CMR results to evaluate for LVMI, $172$ remained after a pre-processing step, $106$ of which are continuous variables and $66$ categorical variables for $91$ patients. As mention in Section I, the goal is to find the most important risk factors that affect LVMI. 


In this precision medicine case study, we used package H2O in R and applied our approach for $150$ different deep stacked auto-encoders. At first we used random forest method for feature selection and supervised learning. We used Mean Squared Error (MSE) as a measure for regression evaluation. In each iteration, random forest runs for a different number of trees. We compared the average of MSE in different runs based on each deep architecture.

\begin{figure}[H]
 \centering
  \includegraphics[scale= 0.15]{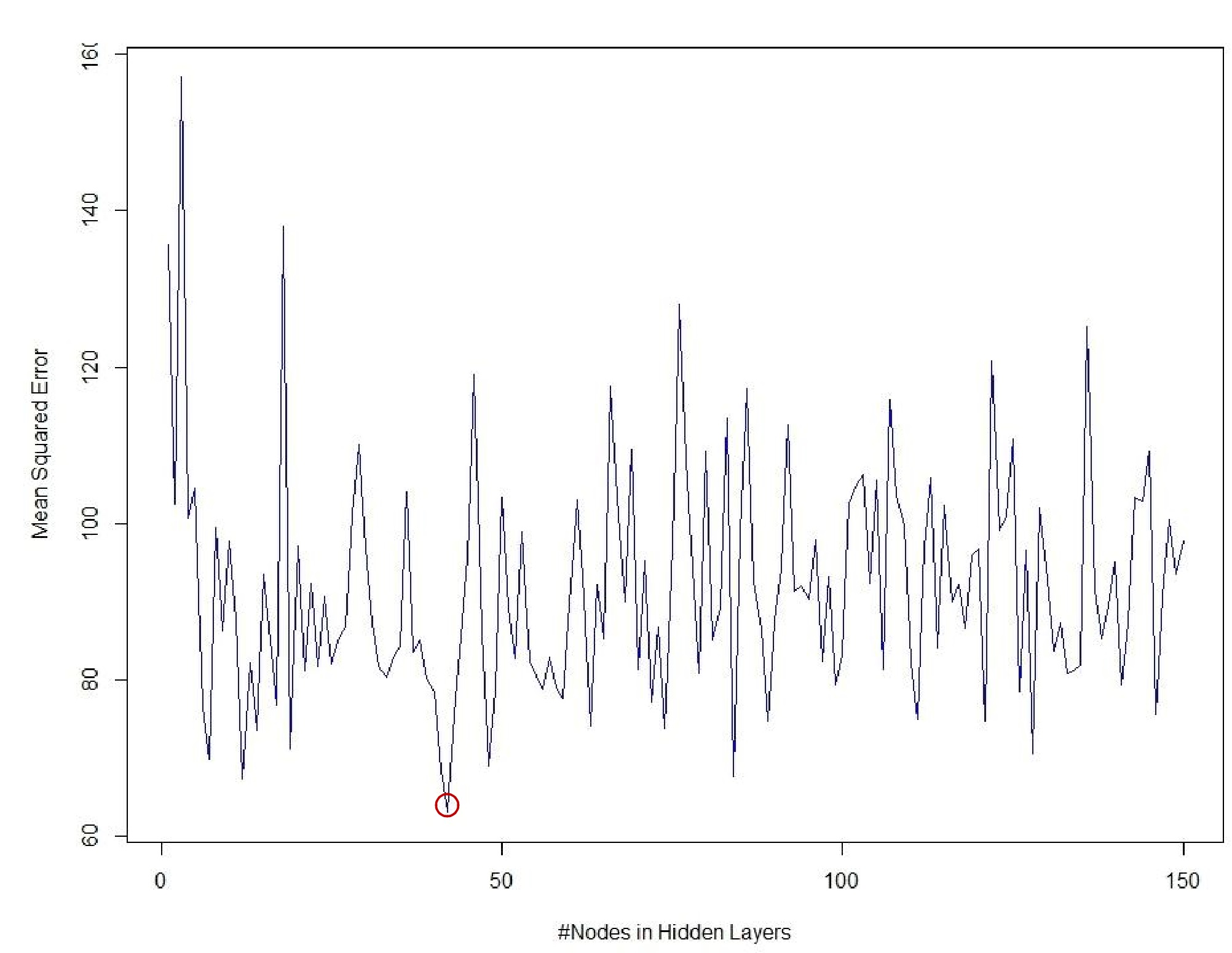}
 \caption{Model Performance using Different Deep Architectures (stacked auto-encoders with different numbers of nodes in hidden layers)}\label{Fig5}
\end{figure}

The result shows that different deep architectures, by changing the number of nodes in the hidden layers, demonstrate different performance. The representation learning produces different data; therefore it leads to different performance. According to Figure \ref{Fig5}, the deep stacked auto-encoders with $n=42$ nodes in second and forth hidden layers yields least error (MSE= $63.93$) among all different architectures. 

\begin{figure}[H]
 \centering
  \includegraphics[scale= 0.25]{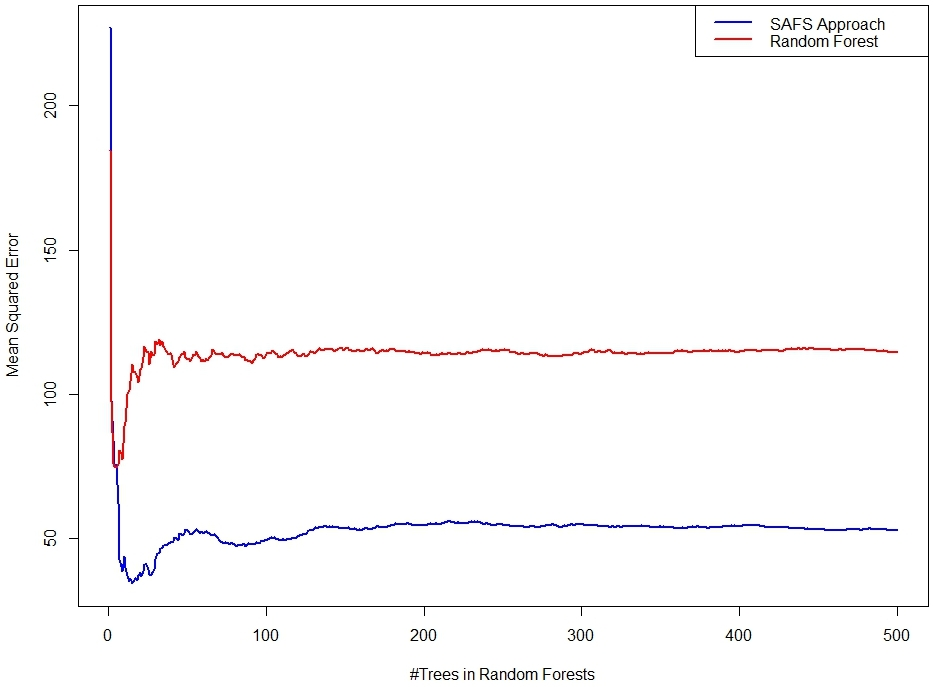}
 \caption{Performance Comparison of SAFS and Random Forest}\label{Fig6}
\end{figure}

By selecting the deep architecture with $n=42$, we ran SAFS approach and then compared it with the results that were generated by random forest method with un-represented input data based on different number of trees. The performance comparison was shown in Figure \ref{Fig6}. It is clear that our SAFS approach achieves a better accuracy (less error) for a different numbers of trees. 

Since the LASSO based approaches are also popular and effective in feature selection from high dimension data, in the second attempt to examine the performance of our approach, we used LASSO for feature selection and prediction in SAFS method, which is implemented in R, by using package “glmnet”. The result shows that the best architecture with least MSE is achieved for $n= 74$ nodes in second and forth hidden layers of stacked auto-encoders (MSE = $361.31$). Figure 7 describes the accuracy comparison of SAFS approach and LASSO based on different amount of Lambda ($\lambda$) parameter, where $\lambda$ is a tuning parameter to determine the number of features to be selected.  It is obvious that SAFS approach has better performance for a wide range of $\lambda$ values, representing many scenarios of feature selections.  

\begin{figure}[H]
 \centering
  \includegraphics[scale= 0.35]{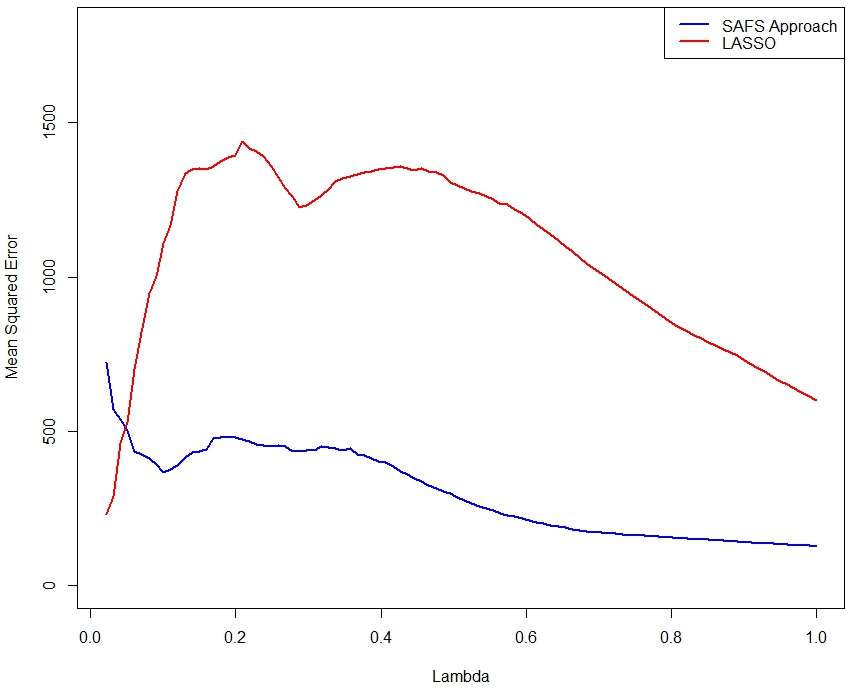}
 \caption{Performance Comparison of SAFS and LASSO}
\end{figure}

The results concluded from both implementations show that using SAFS approach and representation learning significantly outperform random forest and LASSO methods with unrepresented input data and also using random forest as a non-linear classifier in SAFS approach leads to better performance. The SAFS could be an appropriate choice for the precision medicine problem of identifying which patients at risk should be referred for more extensive testing and, which patients with increased LVMI are more or less likely to benefit from treatments.

According to our results obtained from SAFS approach using random forest, the top $15$ significant risk factors that affect LVMI are mentioned in Table I: 

\begin {table}[H]
\small
\caption {Significant Risk Factors Affecting LVMI}\label{table1} 

\centering

\begin{tabular}{| c | l | c |}
\hline\hline
\ Row &  Feature Description & Weight\\
\hline
1 & Troponin Levels & 8.85\\
\hline
2 & Waist Circumference Levels  & 8.07\\
\hline
3 & Plasma Aldosterone  & 7.81\\
\hline
4 & Average Weight  & 7.66\\
\hline
5 & Average Systolic Blood Pressure  & 5.32\\
\hline
6 & Serum eGFR Levels  & 5.21\\
\hline
7 & Total Prescribed Therapeutic Intensity Score  & 4.79\\
\hline
8 & Pulse Wave Velocity  & 4.66\\
\hline
9 & Ejection Duration  & 4.46\\
\hline
10 & Cornell Product  & 4.06\\
\hline
11 & Average Percent Compliance Per Pill Count  & 2.99\\
\hline
12 & Plasma Renin Levels  & 2.86\\
\hline
13 & Transformed Scale Physical Score  & 2.64\\
\hline
14 & Cholesterol Levels  & 2.53\\
\hline
15 & Transformed Scale Mental Score  & 2.43\\
\hline\hline

\end{tabular}
\end {table}

While some of these risk factors represent data specific to the study design, all are plausibly linked with heart risk, representing potential physiological (numbers $1$, $2$, $3$, $4$, $5$, $6$, $8$, $9$, $10$, $12$, and $14$), behavioral (numbers $7$ and $11$), or psychosocial (numbers $13$ and $15$) markers or contributors. In future applications, such an approach to data analysis could be used to make better decisions based on personalized understanding of a specific target subgroup.  

\section{Conclusion}
In this paper, a novel feature selection method called SAFS based on deep learning has been developed. The overall proposed approach includes three phases: representation learning, feature selection and supervised learning. For representation learning, we used stacked auto-encoders with $3$ hidden layers in an iterative learning process that chooses the best architecture for feature selection. We applied our model in a specific precision medicine problem with goal of finding the most important risk factors that affect LVMI, an indicator of cardiovascular disease risk, in African-Americans with hypertension. The results show that our method (SAFS) outperforms random forests and LASSO significantly in feature ranking and selection. In future work we will evaluate our model on related precision medicine problems aimed at reducing cardiovascular disparities and apply different deep architectures as well as stacked auto-encoders for representation learning. 



\begin{thebibliography}{1}

\bibitem{IEEEhowto:1}
J. Hu, A. Perer, and F. Wang, \emph{Data Driven Analytics for Personalized Healthcare},\hskip 1em plus 0.5em minus 0.4em\relax Healthcare Informatics, pp. 529-551, 2016. 

\bibitem{IEEEhowto:2}
W. K Redekop, D. Mladsi, and K. Oka, \emph{The Faces of Personalized Medicine: A Framework for Understanding Its Meaning and Scope},\hskip 1em plus 0.5em minus 0.4em\relax Value in Health, vol. 16, pp. 4-9, 2013.

\bibitem{IEEEhowto:3}
Y. Bengio, A. Courville, and P. Vincent, \emph{Representation Learning: A Review and New Perspectives},\hskip 1em plus 0.5em minus 0.4em\relax IEEE Transactions on Pattern Analysis and Machine Intelligence , vol. 35, pp. 1798-1828, 2013.

\bibitem{IEEEhowto:4}
H. Larochelle, Y. Bengio, J. Louradour, and P. Lamblin, \emph{Exploring Strategies for Training Deep Neural Networks},\hskip 1em plus 0.5em minus 0.4em\relax Journal of Machine Learning Research, vol. 1, pp. 1-40, 2009. 

\bibitem{IEEEhowto:5}
Y. Bengio, \emph{Learning Deep Architectures for AI},\hskip 1em plus 0.5em minus 0.4em\relax Foundations and Trends Machine Learning, vol.1, pp.1-127, 2009. 

\bibitem{IEEEhowto:6}
P. Mamoshina, A. Vieira, E. Putin, and A. Zhavoronkov, \emph{Applications of Deep Learning in Biomedicine},\hskip 1em plus 0.5em minus 0.4em\relax Molecular Pharmaceutics, vol. 13, pp. 1445-1454, 2016.

\bibitem{IEEEhowto:7}
Y. LeCun, Y. Bengio, and G. Hinton, \emph{Deep Learning},\hskip 1em plus 0.5em minus 0.4em\relax Nature, vol. 52, pp. 436-444, 2915.

\bibitem{IEEEhowto:8}
LISA Lab: Deep learning tutorials,\hskip 1em plus 0.5em minus 0.4em\relax http://deeplearning.net/tutorial2.

\bibitem{IEEEhowto:9}
Stanford University: Deep learning tutorials,\hskip 1em plus 0.5em minus 0.4em\relax http://ufldl.stanford.edu/.

\bibitem{IEEEhowto:10}
Y. Bengio, P. Lamblin, D. Popovici, H. Larochelle, \emph{Greedy Layer-Wise Training of Deep Networks},\hskip 1em plus 0.5em minus 0.4em\relax Advances in Neural Information Processing Systems, vol. 19,pp. 153-160,  MIT Press, 2007.

\bibitem{IEEEhowto:11}
S. Zhou, Q. Chen, and X. Wang, \emph{Active Deep Learning Method for Semi-Supervised Sentiment Classification},\hskip 1em plus 0.5em minus 0.4em\relax Neurocomputing journal, vol. 120, pp. 536-546, 2013.

\bibitem{IEEEhowto:12}
Y. Li, C.Y. Chen, and W. Wasserman, \emph{Deep Feature Selection: Theory And Application To Identify Enhancers And Promoters},\hskip 1em plus 0.5em minus 0.4em\relax Journal of Computational Biology , vol. 23, pp. 322-336, 2016.

\bibitem{IEEEhowto:13}
P. Ruangkanokmas, T. Achalakul, and K. Akkarajitsakul, \emph{Deep Belief Networks With Feature Selection For Sentiment Classification},\hskip 1em plus 0.5em minus 0.4em\relax 7th International Conference on Intelligent Systems, Modeling and Simulation, 2016.

\bibitem{IEEEhowto:14}
Q. Zou, T. Zhang, and Q. Wang, \emph{Deep Learning Based Feature Selection For Remote Sensing Scene Classification},\hskip 1em plus 0.5em minus 0.4em\relax IEEE Geoscience And Remote Sensing Letters, Vol. 12, 2015.

\bibitem{IEEEhowto:15}
R. Ibrahim, N.A. Yousri, M.A. Ismail, and N.M. El-Makky, \emph{Multi-Level Gene/Mirna Feature Selection Using Deep Belief Nets And Active Learning},\hskip 1em plus 0.5em minus 0.4em\relax 36th Annual International Conference of the IEEE Engineering in Medicine and Biology Society , 2014.

\bibitem{IEEEhowto:16}
R. Fakoor, f. Ladhak, A. Nazi, and M. Huber, \emph{Using Deep Learning To Enhance Cancer Diagnosis And Classification},\hskip 1em plus 0.5em minus 0.4em\relax 30th International Conference on Machine Learning, 2013. 

\bibitem{IEEEhowto:17}
D. Kumar, A. Wong, and D.A. Clausi, \emph{Lung Nodule Classification Using Deep Features In Ct Images},\hskip 1em plus 0.5em minus 0.4em\relax 12th conference of Computer and Robot Vision (CRV), 2015. 

\bibitem{IEEEhowto:18}
G. Hinton, S. Osindero, Y.W. Teh, \emph{A Fast Learning Algorithm For Deep Belief Nets},\hskip 1em plus 0.5em minus 0.4em\relax Neural Computation, vol. 18, pp. 1527-1554, 2006. 

\bibitem{IEEEhowto:19}
L. Deng, D. Yu, \emph{Deep Learning Mehtods And Applications},\hskip 1em plus 0.5em minus 0.4em\relax Foundations and Trends in Signal Processing, vol. 7. pp. 197-387, 2014.








\end{thebibliography}
\end{document}